\documentclass{article}
\usepackage[preprint]{spconf}
\usepackage{amsmath,amssymb,graphicx,booktabs,hyperref}
\usepackage{subfiles,balance,enumitem}
\hypersetup{hidelinks}
\makeatletter
\renewcommand{\title}[1]{\gdef\@title{\MakeUppercase{#1}}}
\makeatother

\title{From HL to H+L-1 Parameters: A Hankel--Toeplitz Forecaster for Long-Term Time Series Forecasting}
\name{Chaoqi Zhang, Yu Wang, and Haixu Tang
}
\address{Indiana University Bloomington, IN, USA}

\copyrightnotice{%
\parbox[b]{0.48\textwidth}{%
\footnotesize
This work has been submitted to the IEEE for possible publication.
Copyright may be transferred without notice, after which this version may no longer be accessible.}}

\begin{document}
%\ninept
%
\maketitle
\begin{abstract}
Linear forecasters have shown competitive accuracy against Transformer-based models in long-term time series forecasting. We study how classical stationary prediction theory can guide parameter sharing for more compact linear forecasters. For centered second-order stationary processes with nonsingular history covariance, the minimum-MSE finite-window linear predictor factors into a Hankel cross-covariance matrix and an inverse Toeplitz covariance matrix. Shared lags and scale cancellation specify this predictor using $H+L-1$ autocorrelations for lookback $L$ and horizon $H$. Building on the innovations representation, our Hankel--Toeplitz Forecaster (HTF) learns one impulse response that defines both an inverse filter and a forecast map. We characterize the finite-history correction and, under summability assumptions, bound the excess risk of truncating the true filters. HTF uses $H+L-1$ trainable coefficients while allowing a full-rank forecasting matrix. Across seven benchmarks at $L=336$, its horizon-averaged MSE is within 1.2\% of Dense Linear on each dataset with 75--229 times fewer trainable parameters.
\end{abstract}

\section{Introduction}
\label{sec:intro}
Long-term time series forecasting (LTSF) supports planning in energy, transportation, and environmental monitoring. Transformer models for forecasting many steps ahead include Informer \cite{zhou2021informer}, Autoformer \cite{wu2021autoformer}, and PatchTST \cite{nie2023patchtst}. Meanwhile, DLinear demonstrated that simple linear models achieve competitive forecasting accuracy on common benchmarks \cite{zeng2023}, motivating closer study of linear prediction structure.

Compact linear forecasters reduce learned coefficients through frequency-domain interpolation in FITS \cite{xu2024fits}, periodic structure in SparseTSF \cite{lin2024sparse}, temporal and frequency operations in DiPE-Linear \cite{zhao2026dipe}, and segment-level bases in TimeBase \cite{huang2025timebase}. MixLinear combines segment-based trends with low-rank spectral filtering \cite{ma2026mixlinear}. These designs raise a question about the prediction problem itself: what structure does stationary linear prediction impose on a forecasting matrix, and can it guide a compact, interpretable forecaster?

Linear-model analyses characterize architecture equivalences \cite{toner2024} and affine mappings in periodic prediction \cite{li2026affine}. Classical filter design connects prediction objectives to causal filters \cite{wildi2016realtime}; the Wiener--Hopf equations link covariance and innovations descriptions \cite{subbarao2023wienerhopf}. PULSE also invokes Wold decomposition, but models evolving structure and stochastic residuals for nonstationary forecasting \cite{liu2026pulse}. Our focus is coefficient sharing within a linear map.

\begin{figure*}[t]
\centering
\includegraphics[width=\textwidth]{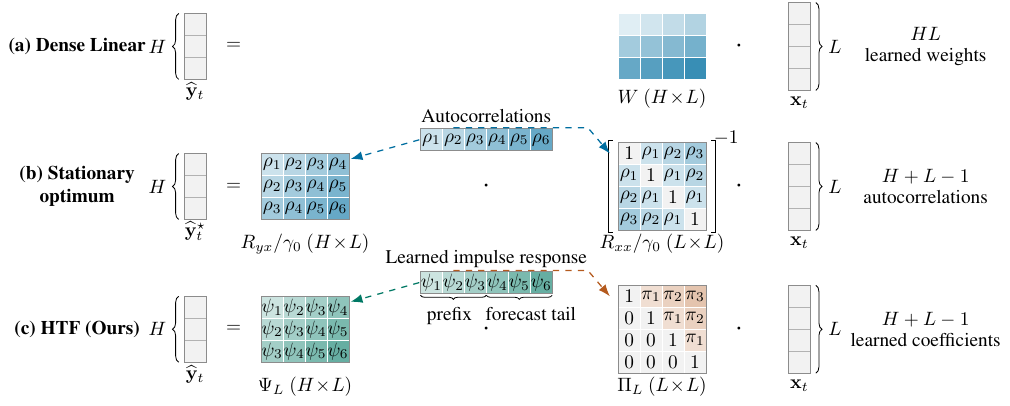}
\caption{Prediction structure ($L=4$, $H=3$), with most-recent-first inputs. Dashed arrows show shared-parameter construction; the inverse filter uses only the prefix. Panels (a,c) show layers before normalization; (b) is the exact stationary predictor.}
\label{fig:overview}
\end{figure*}

Classical Wiener--Hopf theory gives the minimum-MSE finite-window linear predictor for a second-order stationary process as a Hankel cross-covariance matrix multiplied by an inverse Toeplitz covariance matrix \cite{subbarao2023wienerhopf}. Their common scale cancels, leaving $H+L-1$ autocorrelations for lookback $L$ and horizon $H$, versus $HL$ independent weights. Covariances determine prediction coefficients from second-order statistics; innovations isolate information not linearly predictable from earlier history. The latter view suggests inverse-filtering the history, then propagating past innovations forward. We construct both operations from one learned impulse-response sequence, yielding the \emph{Hankel--Toeplitz Forecaster (HTF)} (Fig.~\ref{fig:overview}). We characterize the finite-history boundary correction and bound the excess prediction risk incurred by truncating the true filters.

We evaluate whether this theory-derived parameter sharing preserves forecasting accuracy at a substantially smaller coefficient budget. Across seven LTSF benchmarks at $L=336$, HTF achieves horizon-averaged MSE within $1.2\%$ of Dense Linear on each dataset with $75$--$229\times$ fewer trainable parameters. Our contributions are:
\begin{itemize}[
  topsep=3pt,
  partopsep=0pt,
  itemsep=2pt,
  parsep=0pt,
  leftmargin=12pt
]
\item \textbf{Prediction-theoretic foundation.} We connect the classical finite-window covariance predictor to innovations filtering, derive the finite-history boundary correction, and bound the excess prediction risk of truncating the true filters.
\item \textbf{A theory-derived, interpretable forecaster.} HTF uses one learned impulse-response sequence to construct an inverse filter and a forecast map, requiring $H+L-1$ coefficients while permitting a full-rank forecasting matrix.
\item \textbf{Accuracy--parameter tradeoff.} HTF approaches Dense Linear's accuracy with fewer coefficients, achieving lower mean relative MSE than the evaluated covariance and low-rank controls on their respective grids.
\end{itemize}

\section{Prediction-Theoretic Structure}
\label{sec:theory}
\noindent\textbf{Finite-window prediction.}
Let $x_t$ be a centered, real, second-order stationary process (constant mean and covariance depending only on lag), with autocovariance $\gamma_k=\mathbb E[x_tx_{t-k}]$. Define the most-recent-first history and future by
\begin{equation}
\mathbf x_t=(x_t,\ldots,x_{t-L+1})^\top,\quad
\mathbf y_t=(x_{t+1},\ldots,x_{t+H})^\top.
\end{equation} Assume $R_{xx}=\mathbb E[\mathbf x_t\mathbf x_t^\top]\succ0$.

\noindent\textbf{Classical covariance characterization.}
The matrix minimizing $\mathbb E\|\mathbf y_t-W\mathbf x_t\|_2^2$ satisfies the Wiener--Hopf equations \cite{subbarao2023wienerhopf}:
\begin{equation}
\begin{split}
 W_L^\star&=R_{yx}R_{xx}^{-1},\\
 (R_{xx})_{ij}&=\gamma_{|i-j|},\qquad
 (R_{yx})_{hj}=\gamma_{h+j-1},
\end{split}
\label{eq:wh}
\end{equation}
where $R_{yx}=\mathbb E[\mathbf y_t\mathbf x_t^\top]$, $1\le i,j\le L$, and $1\le h\le H$. Orthogonality of the optimal residual to the history gives $W_L^\star R_{xx}=R_{yx}$; stationarity makes $R_{xx}$ Toeplitz, with constant diagonals, and $R_{yx}$ Hankel, with constant anti-diagonals.

These repeated entries use only lags $0,\ldots,H+L-1$. Dividing both matrices by $\gamma_0$ cancels their scale. The autocorrelations $\rho_k=\gamma_k/\gamma_0$, $k=1,\ldots,H+L-1$, therefore suffice to specify the population optimum, provided they admit a valid stationary covariance extension. This is a sufficient parameterization, not a minimality claim for arbitrary forecasting tasks. For a channel-shared layer minimizing summed stationary quadratic risks, the same argument uses pooled autocovariances.

\begin{table*}[t]
\centering
\caption{Test MSE and trainable parameters (K = thousands), averaged over four horizons at $L=336$. Bold marks the lowest value for each metric in each row.}
\label{tab:accuracy}
\small
\setlength{\tabcolsep}{3pt}
\begin{tabular}{lrrrrrrrrrrrrrr}
\toprule
Dataset & \multicolumn{2}{c}{Dense Linear} & \multicolumn{2}{c}{DLinear} & \multicolumn{2}{c}{FITS} & \multicolumn{2}{c}{SparseTSF} & \multicolumn{2}{c}{DiPE-Linear} & \multicolumn{2}{c}{TimeBase} & \multicolumn{2}{c}{HTF (Ours)} \\
\cmidrule(lr){2-3}\cmidrule(lr){4-5}\cmidrule(lr){6-7}\cmidrule(lr){8-9}\cmidrule(lr){10-11}\cmidrule(lr){12-13}\cmidrule(lr){14-15}
 & MSE & K & MSE & K & MSE & K & MSE & K & MSE & K & MSE & K & MSE & K \\
\midrule
ETTh1 & 0.4088 & 112.90 & 0.4139 & 226.46 & 0.4112 & 40.35 & \textbf{0.4044} & 0.22 & 0.4053 & 1.87 & 0.4093 & \textbf{0.19} & 0.4075 & 0.67 \\
ETTh2 & 0.3429 & 112.90 & 0.3436 & 226.46 & \textbf{0.3388} & 40.35 & 0.3475 & 0.22 & 0.3441 & 5.64 & 0.3535 & \textbf{0.19} & 0.3390 & 0.67 \\
ETTm1 & 0.3614 & 112.90 & 0.3624 & 226.46 & 0.3606 & 16.99 & 0.3834 & 0.11 & 0.3635 & 1.87 & 0.3724 & \textbf{0.06} & \textbf{0.3595} & 0.67 \\
ETTm2 & 0.2581 & 112.90 & 0.2581 & 226.46 & \textbf{0.2568} & 17.62 & 0.2842 & 0.11 & 0.2582 & 1.87 & 0.2940 & \textbf{0.06} & 0.2572 & 0.67 \\
Weather & 0.2481 & 112.90 & 0.2477 & 226.46 & 0.2483 & 7.80 & 0.2769 & 0.15 & \textbf{0.2249} & 7.58 & 0.2935 & \textbf{0.04} & 0.2491 & 0.67 \\
Electricity & 0.1692 & 112.90 & 0.1696 & 226.46 & 0.1695 & 102.88 & 0.1727 & 0.22 & \textbf{0.1655} & 8.78 & 0.1761 & \textbf{0.19} & 0.1705 & 0.67 \\
Traffic & \textbf{0.4333} & 112.90 & 0.4340 & 226.46 & 0.4339 & 102.88 & 0.4360 & 0.22 & 0.4354 & 10.94 & 0.4445 & \textbf{0.19} & 0.4347 & 0.67 \\
\bottomrule
\end{tabular}
\end{table*}

\noindent\textbf{Innovations and the finite-history boundary.}
Wold's representation expresses a purely nondeterministic process (with no perfectly predictable component) through innovations \cite{brockwell1991}. Beyond Wold's assumptions, we require invertibility and absolute summability of both coefficient sequences for the inverse representation and boundary analysis:
\begin{equation}
 x_t=\sum_{j\ge0}\psi_j\varepsilon_{t-j},\quad
 \varepsilon_t=\sum_{k\ge0}\pi_kx_{t-k},\quad \psi_0=\pi_0=1.
 \label{eq:wold}
\end{equation}
An innovation $\varepsilon_t$ is the part of the current observation that cannot be linearly predicted from the past. The impulse-response coefficient $\psi_j$ describes its contribution $j$ steps later; the inverse-filter coefficients $\pi_k$ recover innovations from observed history. Let $P_t$ project onto the entire observed past. Future innovations are orthogonal to the past, while past innovations affect future observations through the impulse response. Thus $P_tx_{t+h}=\sum_{i\ge0}\psi_{h+i}\varepsilon_{t-i}$.
Expanding the two filters in observed samples gives the lag-$k$ weight $q_{hk}$ for step $h$ and the outside-window contribution $u_{h,L}$:
\begin{equation}
 q_{hk}=\sum_{i=0}^{k}\psi_{h+i}\pi_{k-i},\qquad
 u_{h,L}=\sum_{k\ge L}q_{hk}x_{t-k}.
\end{equation}
Let $(A_L)_{hj}=q_{h,j-1}$ for $1\le j\le L$, and write $\mathbf u_L=(u_{1,L},\ldots,u_{H,L})^\top$.

\noindent\textbf{Proposition 1 (boundary correction).}
Under these summability assumptions, the true filters in~\eqref{eq:wold} satisfy
\begin{equation}
 W_L^\star=A_L+\mathbb E[\mathbf u_L\mathbf x_t^\top]R_{xx}^{-1}.
 \label{eq:boundary}
\end{equation}
For $\mathcal R(W)=\mathbb E\|\mathbf y_t-W\mathbf x_t\|_2^2$, the following bound vanishes as $L\to\infty$ for fixed $H$:
\begin{equation}
 0\le\mathcal R(A_L)-\mathcal R(W_L^\star)
 \le\gamma_0\sum_{h=1}^{H}\Big(\sum_{k\ge L}|q_{hk}|\Big)^2
 \label{eq:tail}
\end{equation}
\emph{Proof.} Expanding the two filters yields $P_t\mathbf y_t=A_L\mathbf x_t+\mathbf u_L$. Projection onto the finite history gives~\eqref{eq:boundary}. The excess risk equals the squared $L^2$ norm of the projection of $\mathbf u_L$. Projection is contractive and $\|x_t\|_{L^2}=\sqrt{\gamma_0}$, giving~\eqref{eq:tail} by the triangle inequality. Absolute summability makes each convolution tail vanish. $\square$

Correlations between earlier history and the visible window produce the correction omitted by truncation. The bound concerns true filters, not learned coefficients. Finite-section approximation is classical \cite{finitesection2023}; the identity clarifies the finite construction.

\section{Hankel--Toeplitz Forecaster}
\label{sec:model}
HTF learns $\theta=(\psi_1,\ldots,\psi_{H+L-1})$, with $\psi_0=\pi_0=1$. It derives the inverse-filter prefix through
\begin{equation}
 \pi_n=-\sum_{m=1}^{n}\psi_m\pi_{n-m},\qquad 1\le n<L.
 \label{eq:inverse}
\end{equation}
Define the inverse-filter matrix $\Pi_L\in\mathbb R^{L\times L}$ and Hankel forecast map $\Psi_L\in\mathbb R^{H\times L}$:
\begin{equation}
\begin{split}
 (\Psi_L)_{hj}&=\psi_{h+j-1},\quad
 (\Pi_L)_{ij}=\begin{cases}\pi_{j-i},&j\ge i,\\0,&j<i,\end{cases}\\
 \widehat{\mathbf y}_t&=W_\theta\mathbf x_t,\qquad W_\theta=\Psi_L\Pi_L.
\end{split}
\label{eq:model}
\end{equation}
The filtered history $\Pi_L\mathbf x_t$ estimates past innovations, whose effects $\Psi_L$ propagates forward using all learned coefficients. The prefix $\psi_{1:L-1}$ determines the inverse filter, while the tail $\psi_{L:H+L-1}$ supplies the remaining forecast coefficients. Sharing gives $H+L-1$ trainable coefficients. The true filters yield $A_L$ in Proposition~1; learned filters instead provide a modeling prior, without necessarily recovering white innovations. Chronological implementation inputs are reversed to match our coordinates. The finite filters have bounded-input, bounded-output stability; the recursion imposes no guarantee of a stable infinite extension or globally valid Wold spectral factor.

\noindent\textbf{Full-rank capacity.}
Since $\Pi_L$ is unit triangular, $\operatorname{rank}(W_\theta)=\operatorname{rank}(\Psi_L)$. For $r=\min(H,L)$, setting $\psi_r\ne0$ and the other learned coefficients to zero gives an invertible anti-diagonal $r\times r$ minor. Thus HTF shares coefficients while permitting full rank before normalization.

\noindent\textbf{Initialization and normalization.}
Yule--Walker equations fit an autoregression from training-window autocorrelations; its impulse response initializes $\psi$ \cite{brockwell1991}. Forecast MSE then updates the shared sequence. One layer is shared across channels. HTF and Dense Linear use zero forecast bias and symmetric per-window normalization inspired by RevIN \cite{kim2022revin}, without learned affine normalization parameters. Subtracting the input mean $\mu$, dividing by its scale, and undoing normalization cancels that scale:
\begin{equation}
 \widehat{\mathbf y}=W(\mathbf x-\mu\mathbf1_L)+\mu\mathbf1_H,
 \label{eq:normalization}
\end{equation}
where $\mathbf1_d$ is the length-$d$ all-ones vector. The raw-space map is linear and preserves constants, but its trained coordinate matrix $W$ is not asserted to equal~\eqref{eq:wh}. HTF's savings concern learned parameters; cached dense inference still costs $O(HL)$ per sample and channel.

\section{Experiments}
\label{sec:experiment}
\noindent\textbf{Datasets and protocol.}
We evaluate seven LTSF benchmarks \cite{zhou2021informer,zeng2023}: ETTh1/2, ETTm1/2, Weather, Electricity, and Traffic. Lookback $L=336$ covers all seven; $L=720$ covers the four ETT datasets, each with $H\in\{96,192,336,720\}$. ETT splits are 12/4/4 months; others use chronological 70/10/20\% splits. Channel standardization uses training statistics. Models share window normalization, MSE loss and evaluation, with architecture-specific settings. Adam uses batches of 256, at most 100 epochs, patience 10, and the best validation-MSE checkpoint. HTF and Dense Linear use learning rates $10^{-3}$ and $5\times10^{-4}$; both use zero forecast bias. Other architectures retain their bias terms. Parameter counts include the dataset-specific configurations evaluated here. Baselines are common-framework reimplementations: DiPE-Linear follows its official implementation without its auxiliary loss; TimeBase's original normalization and regularization are not retained.

\begin{table}[t]
\centering
\small
\caption{Zero-bias internal controls, $L=336$. Within each block, differences are mean relative MSE against Dense Linear (negative is better). Counts are fitted coefficients; COV-PACF denotes the covariance control parameterized by partial autocorrelations.}
\label{tab:rank}
\begin{tabular}{lrr}
\toprule
Model & Coefficients & Difference \\
\midrule
\multicolumn{3}{l}{All 28 settings, $H\in\{96,192,336,720\}$} \\
HTF (Ours) & $H+L-1$ & $-0.12\%$ \\
Ridge (zero bias) & $HL$ & $-0.61\%$ \\
Ridge-projected, rank 1 & $H+L$ & $+80.45\%$ \\
Ridge-projected, rank 2 & $2(H+L)$ & $+39.19\%$ \\
Ridge-projected, rank 8 & $8(H+L)$ & $+10.69\%$ \\
\midrule
\multicolumn{3}{l}{Matched 14 settings, $H\in\{96,336\}$} \\
HTF (Ours) & $H+L-1$ & $-0.11\%$ \\
COV-PACF & $H+L-1$ & $+1.43\%$ \\
\bottomrule
\end{tabular}
\end{table}

\begin{figure}[t]
\centering
\includegraphics[width=\columnwidth]{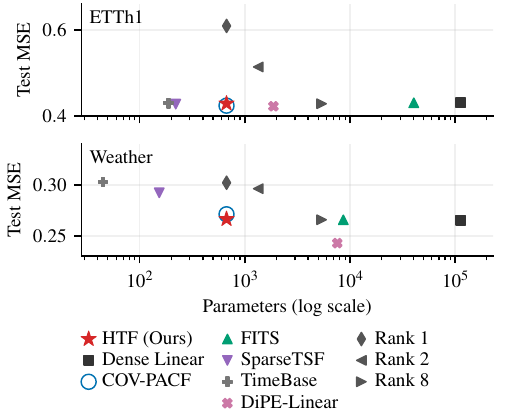}
\caption{Accuracy--parameter tradeoff at $L=H=336$. Task-specific counts and MSE follow our common protocol; lower/left is better. Internal controls use zero bias. Rank 1/2/8 denote Ridge-projected controls, counting factor coefficients.}
\label{fig:tradeoff}
\end{figure}

\noindent\textbf{Accuracy and parameter efficiency.}
HTF's dataset-level mean MSE is within $1.2\%$ of Dense Linear on each of the seven datasets and lower on four (Table~\ref{tab:accuracy}). Across 28 settings at $L=336$, its mean relative MSE difference is $-0.12\%$, with $75$--$229\times$ compression. Here we average $100(\mathrm{MSE}_{\mathrm{HTF}}/\mathrm{MSE}_{\mathrm{Dense}}-1)$ across settings; this differs from comparing horizon-averaged dataset errors. The 16 ETT settings at $L=720$ give $-1.42\%$ with $85$--$360\times$ fewer parameters.

Three paired repetitions cover 20 settings: all four horizons on ETTh2, ETTm2, Weather, Electricity, and Traffic at $L=336$. The mean relative MSE difference on this fixed grid is $+0.01\%$, with an observed range of $-0.07\%$ to $+0.09\%$ across paired repetitions. These observations do not establish statistical equivalence.

Against compact linear forecasters, HTF's mean relative MSE is $0.01\%$ below FITS and $1.77\%$ above DiPE-Linear at $L=336$. SparseTSF and TimeBase are smaller, while HTF has $4.56\%$ and $6.31\%$ lower mean relative MSE in our pipeline. On the 16 ETT settings at $L=720$, HTF is $1.24\%$ below DiPE-Linear and $0.70\%$ below FITS. At nearly the same coefficient budget, HTF is more accurate than Ridge-projected rank-1 compression and approaches Dense Linear on both tasks (Fig.~\ref{fig:tradeoff}).

In a separate Weather ablation with learned biases at $L=H=336$, per-channel HTF and Dense Linear both reduce MSE by about $10\%$, using $21\times$ their shared-model parameter counts. This supports channel-specific filtering, without attributing the benefit uniquely to HTF.

\noindent\textbf{Sharing versus rank restriction.}
We compare HTF with a \emph{Ridge-projected low-rank comparator}, which projects a closed-form Ridge fit $\widehat W$ onto the leading eigenvectors $U_r$ of $\widehat W S_{xx}\widehat W^\top$, where $S_{xx}$ is the centered training input Gram matrix. The resulting map is $U_rU_r^\top\widehat W$. This Ridge-dependent estimator differs from exact reduced-rank regression \cite{izenman1975rrr}.

At $L=H=336$, HTF uses 671 coefficients versus 672 for rank 1. Across the 28 settings, even rank 8 has higher mean relative MSE than HTF.

\noindent\textbf{Covariance and fitting references.}
We evaluate a covariance predictor parameterized by the partial autocorrelation function (COV-PACF). It constructs a positive-definite Toeplitz covariance sequence \cite{ramsey1974pacf} and applies~\eqref{eq:wh}, at HTF's coefficient budget. On 14 matched zero-bias settings (seven datasets, $H=96,336$), mean relative MSE is $+1.43\%$ for COV-PACF and $-0.11\%$ for HTF (Table~\ref{tab:rank}). Differing optimization precludes isolating the forecast tail. Zero-bias Ridge has a mean relative MSE $0.61\%$ below Dense Linear over the 28 settings, providing a closed-form fitting reference.

\noindent\textbf{Signal interpretation.}
Spectral flatness is the geometric-to-arithmetic mean power ratio. For zero-bias HTF at $L=H=336$, median flatness of inverse-filtered nonconstant training windows ranges from $0.209$ to $0.521$, versus $0.565$ for an iid Gaussian reference with matched normalization, cropping and spectral estimation. The inverse-filtered signals are not uniformly white; innovations motivate the architecture without establishing exact innovation recovery.

\section{Conclusion}
HTF translates stationary prediction structure into a shared inverse filter and forecast map. The boundary analysis distinguishes true-filter truncation from the exact finite-window predictor. The resulting $H+L-1$-coefficient parameterization permits full rank and approaches Dense Linear's accuracy on the evaluated benchmarks.

\clearpage
\raggedbottom
\balance
\bibliographystyle{IEEEbib}
\bibliography{refs}
\end{document}